\documentclass{article}
\usepackage{spconf,amsmath,graphicx}

\usepackage{booktabs}
\usepackage{enumitem}
\usepackage{amssymb}

\title{Importance of negative sampling in weak label learning}
\name{Ankit Shah,  Fuyu Tang, Zelin Ye,  Rita Singh, Bhiksha Raj}
	\address{Carnegie Mellon University\\
	5000 Forbes Ave, Pittsburgh, PA 15213}

\begin{document}
\ninept
\maketitle
\begin{abstract}

Weak-label learning is a challenging task that requires learning from data “bags” containing positive and negative instances, but only the bag labels are known. The pool of negative instances is usually larger than positive instances, thus making selecting the most informative negative instance critical for performance. Such a selection strategy for negative instances from each bag is an open problem that has not been well studied for weak-label learning. In this paper, we study several sampling strategies that can measure the usefulness of negative instances for weak-label learning and select them accordingly. We test our method on CIFAR-10 and AudioSet datasets and show that it improves the weak-label classification performance and reduces the computational cost compared to random sampling methods. Our work reveals that negative instances are not all equally irrelevant, and selecting them wisely can benefit weak-label learning.

\end{abstract}
\section{Introduction}
\label{sec:intro}

Machine learning models typically need large, labeled datasets to achieve state-of-the-art performance. This presents a bottleneck: supervised learning approaches do not scale well in real-world scenarios due to a lack of labeled data. Labeling large datasets accurately requires a significant investment of
time and money. Furthermore, skilled annotators with domain expertise are required for the task, and their judgment and biases limit the accuracy and completeness with which they can label data. The approach of \textbf{\textit{learning from weak labels}} has gained popularity in recent research \cite{sun2010multi,shah2018closer,wang2020learning}  due to its lower annotation costs, scalability, real-world applicability, speed, and robustness, etc. to name a few. 


Weakly-labeled learning is a subset of supervised learning where the training data is imperfectly labeled. In the ideal case of supervised learning, each data point is paired with a precise label that indicates the desired output. Weak labels, though imperfectly labeled, are more potent as such data is available at a large scale on the web. These labels may be noisy, incomplete, or come in a form that provides only partial information about the true label.  An example of the curation of the labels is via scraping data from the web to collect the weak information for the labels. In most cases, the weak label implies the knowledge of the presence/absence of an event in a data bag, e.g., an image bag with several pictures or an audio clip, rather than the exact location compared to the strong labels. To gather the weak labels, there is no need to find the exact timestamp of a dog bark's segment in one audio clip or specific class labels for every image in a bag. Without the demand for strongly labeled data, one can easily acquire many weak labels for data samples from external sources/the web. It is worth noting that this is an efficient way to circumvent the annotator's subjective bias when tagging a data sample since reaching a consensus in a weak label setting is not difficult.

In the realm of machine learning with weak labels, it's not uncommon to encounter an imbalance between positive and negative instances. For weak label classification tasks, we categorize the data or "bags" into two primary types \textbf{\textit{Positive}} and \textbf{\textit{Negative}}. Bags containing instances from a specific class such as music - are labeled positive, while those devoid of music are considered negative. Positive examples are often sparse, while negative examples are abundant but often irrelevant to the target class. For instance, in training an audio classifier for e.g. crash sounds with weakly labelled recordings collected from the web, it is fairly challenging to find recordings tagged with the presence of this event. In contrast, enormous numbers of recordings \textit{not} tagged with it can easily be found. The preponderance of negative instances leads to challenges during training, which can unduly influence the classifier's decision-making process.  The makes the choice of training samples, particularly the negative instances, is crucial and a decision often made through particular sampling strategies. The need for improved sampling strategies to select the most distinguishing negative examples from positive ones is evident, especially since this topic has been extensively researched in the context of strongly labeled data but remains underexplored for weak labels. Our research investigates novel sampling methods that modify the treatment of negative instances in scenarios involving weak labels. Utilizing active sampling mechanisms, we focus on choosing negative sets that offer both diversity and informational value, leading to a more equitable and effective training regimen. This research marks a significant departure from traditional practices which often utilizes the random sampling approach, extending the utility of weak label learning in various applications such as audio and image recognition. We show that these new sampling techniques not only surpass conventional random sampling in weak-label scenarios but also enhance performance across a range of categories.

In this paper, we address the problem of selecting optimal sets of negative bags for training with weak labels. Concretely, we address the question on how to optimally select negative bags that contribute meaningfully to training an effective weak classifier? To mitigate this issue, the concept of \textbf{\textit{Negative Sampling}} becomes pivotal. We propose strategies such as Gradient embedding, entropy-based approach, SVM-based margin strategy as well as BADGE strategy for the sampling of negatives. We explore ways to deploy these sampling strategies without meaningful slowdown to the training time while enhancing the model to converge faster through an early stopping mechanism due to a better selection of negative samples. We demonstrate the applicability of  sampling strategy for image and audio classification tasks. The remainder of this paper is organized as follows: Section \ref{sec:Related Work} reviews related work, Section \ref{sec:Notation} introduces the notation used, Section \ref{sec:sampling} details our proposed sampling strategies, and Section \ref{sec:Experiments} presents our experimental setup and results.


\section{Related Work}
\label{sec:Related Work}
\subsection{Weak Label Learning}
\emph{Weak label learning}, or the problem of learning from weakly labelled data, has been the subject of considerable research in recent times \cite{shah2018closer,10.1145/2964284.2964310}. While \cite{10.1145/2964284.2964310} consider the vanilla weak-label problem in the context of audio, more relevant to this paper \cite{shah2018closer} studied how characteristics such as label density, i.e. the fraction of weakly-labelled training instances that must be positive, and label corruption, i.e. noise in the weak labels,  influence training. They also proposed a CNN-based weak-label learning model for large-scale AED. \cite{zhang2021training} introduced semi-weak labels as weak labels with count information and developed a two-stage framework for semi-weak label learning. Since weak labels are generally applied to collections (or ``bags'') of instances, it not only learns classification at bag level but can also be extended to learning instance-level classifiers with counts assigned to bags. However, semi-weak labels are rarely available, especially for audio or video data.

\subsection{Negative Sampling}
Negative sampling has been studied in context of strongly labeled data.
Knowledge graph embedding is a technique used in machine learning to represent structured knowledge in a knowledge graph into continuous vector representations. In \cite{zhang2019nscaching} introduces an approach to knowledge graph embedding that leverages caching to efficiently sample and update negative triplets thereby improving training performance compared to methods using generative adversarial networks (GANs).  \cite{yang2020understanding} studied the influence of negative sampling on objective and risk of embedding learning, and proposed a negative sampling strategy MCNS based on their theory, which approximates the positive distribution with self-contrast approximation and accelerated by Metropolis-Hastings. \cite{ahrabian2020structure} incorporates the richness of graph structure and designed a structure-aware sampling strategy to sample semantically meaningful negative samples.


\subsection{Active Learning}
Active learning strategies often aim to identify the most uncertain or informative samples for labeling. For instance, Ghasemi et al. \cite{ghasemi2011active} proposed an uncertainty-based approach that leverages Bayes rules for positive and unlabelled data, where the probability density function measures the informativeness of samples to select the most uncertain sample. Ma et al. \cite{ma2020active} extended this by introducing a gradient-based uncertainty measure for contrastive learning, inspired by Ash et al. \cite{ash2019deep} solution for the diversity of samples.

Chen et al. \cite{chen2011active} devised a strategy that uses a multi-model committee to assess data's uncertainty and informativeness. Their method gives preference to low-variance minority classes during sampling, aiming to counteract class imbalance in the training set. \cite{shi2019integrating} proposed a Kernel machine Committee-based model that integrates Bayesian (RVM) and discriminative (SVM) kernel machines for multi-class data sampling. It fits the overall data well and can capture the decision boundaries, thus effectively sampling informative data for multi-class training.

Learning-based methods that can help with sampling have also been studied. \cite{shao2019learning} proposed a learning-based active learning framework, where a sampling model and a boosting model mutually learn from each other iteratively to improve performance. Their sampling model incorporates uncertainty sampling and diversity sampling into a unified process for optimization to actively select the most representative and informative samples.

\section{Notation and Setting}
\label{sec:Notation}
We consider the weak label classification problem. Denote by $\mathcal{X}$ the instance set, including $\mathcal{X^+}$ and $\mathcal{X^-}$ which represent positive and negative data instances respectively. Denote by $\mathcal{B}$ the bag set, including $\mathcal{B^+}$ and $\mathcal{B^-}$ which represent positive and negative data bags respectively. Denote by $D_\mathcal{X}$ the instance distribution, including $D_{\mathcal{X}^+}$ representing the positive and $D_{\mathcal{X}^-}$ representing the negative. Denote by $D_\mathcal{B}$ the bag distribution from which our data bags are drawn, including $D_{\mathcal{B}^+}$ representing the positive bag distribution and $D_{\mathcal{B}^-}$ representing the negative. 
Let $K$ be the number of classes, $N_B$ be the number of instances in the bag and $N$ be the number of bags. Define by $\mathcal{Y}$ the label space and $[K]:=\left\{1,2,\cdots,K\right\}$. In a multiclass weak classification problem, $\mathcal{Y}=[K]$. We define a set $\mathcal{S}$ of bags $\left\{(b_0,y_0),(b_1,y_1),\cdots,(b_N,y_N)\right\}$, where each $b_i\sim D_{\mathcal{B}}$ represents a data bag containing a group of instances $\left\{ x^0_i,x^1_i,\cdots,x^{N_B}_i\right\}$, $x_i^j\sim D_\mathcal{X}$ and $y_i$ represents the label for this bag $b_i$. Let's consider the binary weak classification case first, which will also be our focus in this paper: $K = 2$ and $\mathcal{Y}=[2]$. In this scenario,  
\begin{equation}
    y_i=\left\{
\begin{aligned}
1 &, & \exists\ x^j_i \in b_i, \quad x^j_i \sim D_{\mathcal{X}^+}\\
0 &, & \forall\ x^j_i \in b_i, \quad x^j_i \sim D_{\mathcal{X}^-}\\
\end{aligned}
\right.
\end{equation}
denoting the bag $b_i$ as positive or negative ($b_i\sim D_\mathcal{B^+}$ or $b_i\sim D_\mathcal{B^-}$). We consider the negative sampling in weak label learning setup, where the model receives a set of weakly labeled set $\mathcal{S}$ mixed up with all the positive set $\mathcal{S}^+$ and selected negative set $\mathcal{S}^-$ where each $b\in \mathcal{S^-}$ is picked from a negative dataset $\mathcal{N}$ sampled according to $D_\mathcal{B^-}$. 

Given a weak classifier $h: \mathcal{B} \rightarrow \mathcal{Y}$, which maps bags to weak labels, we denote the $0/1$ error of $h$ on $(b,y)$ as $\ell_{01}(h(b),y)=\mathbb{I}(h(x)\neq y)$. The performance of a weak classifier $h$ could be simply measured by its expected $0/1$ error, i.e., $\mathbb{E}_{D_\mathcal{B}}[\ell_{01}(h(b),y)]=\mathrm{Pr}_{(b,y)\sim D_{\mathcal{B}}}(h(b)\neq y)$.  In the context of weak label learning, the objective of negative sampling is to devise a sampling strategy that produces an appropriate set $\mathcal{S^-}$ aimed at minimizing the model's expected $0/1$ error on the dataset.

\section{Sampling Strategies}
\label{sec:sampling}
In this section, we present several widely used sampling strategies applicable for weak labels. We only sample bags from the pool of the negative, meaning $b_i\sim D_\mathcal{B^-}$. The strategies are as follows:-
\begin{enumerate}[leftmargin=12px]  
\item \textbf{Random}: Randomly select $k$ bags with uniform distribution at each epoch. Each negative bag has an equal chance of being selected. This strategy is the baseline for all the sampling strategies.

\item \textbf{Margin}: An uncertainty-based strategy that samples $k$ bags with the lowest margin, which is defined as the difference between positive probability and negative probability in each output result.
\begin{equation}
    S_{M}^- = \underset{b_i \in D_\mathcal{B^-}}{\mathrm{argmin}}\, |P_{\theta}(\hat{y}_{positive}|b_i)-P_{\theta}(\hat{y}_{negative}|b_i)|
\end{equation}

\item \textbf{Entropy}: An uncertainty-based strategy that samples $k$ bags with the highest entropy scores. The entropy score is computed as the entropy of the predicted positive-negative distribution.
\begin{equation}
    S_{E}^- = \underset{b_i \in D_\mathcal{B^-}}{\mathrm{argmax}}\, -\sum P_{\theta}(\hat{y}_{c}|b_i)\log(P_{\theta}(\hat{y}_{c}|b_i))
\end{equation}
\item \textbf{Gradient Embedding}: An uncertainty-based strategy that selects $k$ bags that generates the largest gradient in terms of L2 norm, which is a simplified version of BADGE \cite{ash2019deep} without K-MEANS++\cite{Arthur2007}.
\begin{equation}
g_b = \frac{\partial}{\partial \theta_{\text{out}}} \text{Loss}(f(x; \theta), \hat{y}(b_i))
\end{equation}

\begin{equation}
S_{G}^- = \underset{b_i \in D_\mathcal{B^-}}{\text{argmax}}\, \Vert g_b \Vert_2
\end{equation}


\item \textbf{BADGE}: The strategy combines uncertainty and diversity by picking bags according to gradient embedding as centers of $k$ clusters in the pool. Bag with the largest gradient embedding are selected and K-MEANS++ helps incorporate more diverse samples \cite{ash2019deep}.

  
\end{enumerate}

\section{Experiments and Results}
\label{sec:Experiments}
We first implement random sampling on negative data as a baseline for comparison with every other sampling strategy. We experimented with image "bag" level and weak audio level classification to validate our sampling strategies.

\subsection{Image Bag Classification}
\subsubsection{Dataset}
\vspace{-1mm}
To create a robust negative sampling strategy for weakly-labeled data, we generate image bags from randomly selected images of CIFAR-10 \cite{krizhevsky2009learning} such that instances in the same bag can be irrelevant. CIFAR-10 Dataset contains 60,000 images. We generate bags with fixed size of 5 and uniformly distribute the label density, which denotes the proportion of positive data in positive bags among all the bags. To alleviate the influence caused by the single image being learned unevenly multiple times in a single epoch, each image only occurs in one bag. Although the full dataset is unbalanced for a single class, its coverage and size are essential to the robustness of training. Any subset sampled from the original dataset will likely be at a disadvantage. Therefore, results on a full, weakly labeled dataset naturally become the ideal benchmark we intend to surpass. 

\subsubsection{Network Architecture}
\vspace{-1mm}
 Under the weak label setting, we lack individual image labels and only have bag level labels. To solve this problem, we designed a simple yet effective model structure. First, we use a traditional CNN-based model to classify each image. Given a bag of images $\left\{ x_0, x_1, \cdots, x_{B_N-1} \right\} \in \mathcal{R}^{32\times32\times3}$, the preliminary classification score $P_i \in \mathcal{R}^{K}$ is predicted for each image. Note that the scores here are only an expected distribution for each image in the bag without any strong labels. Then we use an aggregation layer, e.g., mean-pooling or max-pooling following or followed by a Softmax layer, to produce the final classification score in bag level $P_{\mathcal{B}} \in \mathcal{R}^K$ aggregating all the scores above.

For the downstream classification problem, we first choose a Simple CNN model (2 Convolution layers followed by 3 Fully-connected layers) and ResNet18\cite{DBLP:journals/corr/HeZRS15} to examine strategies' effectiveness on different models. Due to overfitting issues with ResNet18, we proceeded with the Simple CNN for all subsequent experiments. In terms of aggregation layers, we explored three configurations: Softmax followed by MeanPooling, MeanPooling followed by Softmax, and Softmax followed by MaxPooling. Our findings indicate that the sequence of MeanPooling followed by Softmax consistently yielded the best performance. This effectiveness is likely attributed to Softmax outputs being bounded between 0 and 1, which, when mean-pooled, enhances the model's ability to discern events within a bag. Conversely, the underperformance of MaxPooling can be ascribed to its sensitivity to noise, causing the model to erroneously assign positive labels based on high scores, irrespective of the true nature of the bag.


\subsubsection{Experiment Setting}
Our experimental setup for image bag classification uses Python version 3.8.11, PyTorch v1.9.0, and CUDA v11.1, running on NVIDIA GeForce GTX 1070 GPU. We investigated the impact of resampling frequency on model performance, specifically contrasting resampling at each epoch against resampling every three epochs. Our findings revealed no significant differences, leading us to opt for epoch-wise resampling in subsequent tests. To comprehensively evaluate the robustness and effectiveness of our proposed sampling strategies, we employed a range of techniques as outlined in Section \ref{sec:sampling}. These methods primarily focus on selecting from the available negative bags to curate a truncated training dataset. Evaluation metrics include Average Precision (AP) and Receiver Operating Characteristic Area Under the Curve (ROC-AUC). By comparing these scores with those from a random sampling baseline, we aim to substantiate the merits of our approach.

\subsubsection{Experiment Result}

Table \ref{tab:results_cifar} shows the results of 2 classes in the CIFAR dataset. In the experiments, the KL-Divergence method is implemented in two ways: one computes KLD based on the embedding extracted from the model, and the other uses output probability distribution. As we can see, Random i.e. our baseline sampling strategy, performs relatively well for test data. The performance with other negative sampling strategies like Gradient Embedding is not consistently as good as the baseline. However, strategies like BADGE, SVM-Margin, and KL-prob can beat random sampling most of the time and achieve AP and AUC asymptotical to the Full dataset.

\begin{table}[ht]
  \caption{\small Comparison between CNN model using different sampling strategies}
  \small
  \label{tab:results_cifar}
 \resizebox*{1.0\columnwidth}{!}{
  \begin{tabular}{ c | c | c | c | c}
  \toprule
{\em Sampling strategy} & {\em Class 0 AP} & {\em Class 0 AUC} & {\em Class 1 AP} & {\em Class 1 AUC}  \\
\hline
 Full Set (No filtering) &            0.88 & 0.95 & 0.93 & 0.97  \\ 
 Random &              0.82 & 0.93 & 0.87 & 0.95  \\ 
 Least Confidence &    0.81 & 0.92 & 0.82 & 0.93 \\
 BADGE &               0.85 & 0.94 & 0.87 & 0.95 \\
 Entropy &             0.81 & 0.92 & 0.82 & 0.93 \\
 KL\_embedding &       0.83 & 0.93 & 0.89 & 0.95 \\
 Grad\_embedding &     0.83 & 0.93 & 0.84 & 0.94 \\
 SVM-Margin &          0.87 & \textbf{0.95} & 0.87 & \textbf{0.95} \vspace{-1mm}\\ 
\bottomrule
  \end{tabular} }

\end{table}






\begin{center}
\begin{figure*}[htb]
\includegraphics[width=1.0\textwidth, height = 5.7cm]{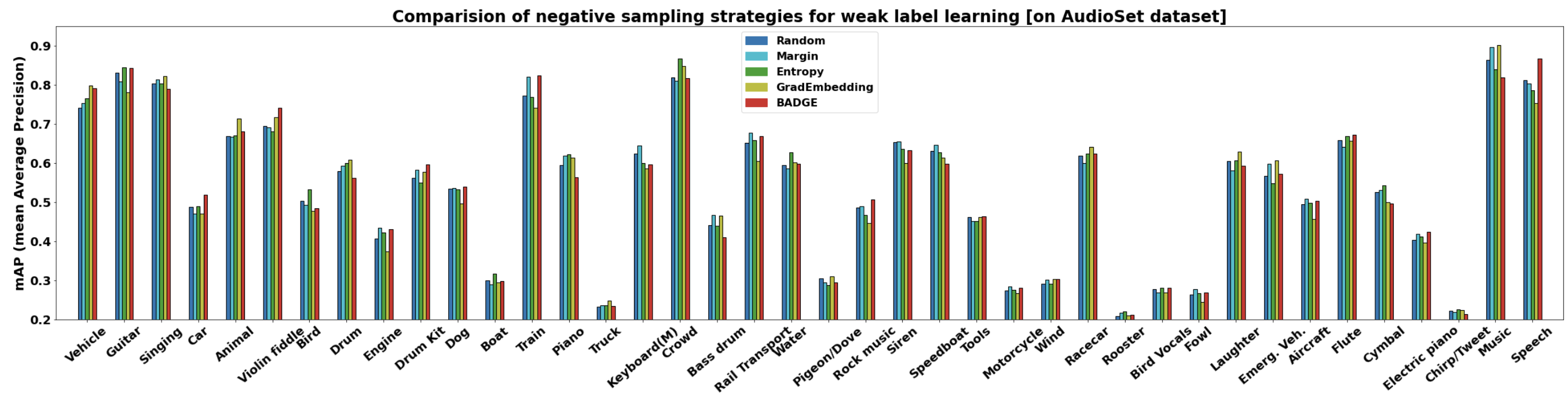}
    \caption{Comparing different negative sampling strategies for weak label learning on AudioSet}
    \label{fig:audioset_result}

\end{figure*}
\end{center}

\vspace*{-\baselineskip}


\subsection{Audio Classification}

\subsubsection{Experiment Setting}
We use AudioSet\cite{gemmeke2017audio}, a large, weakly labeled and multi-class dataset comprising 527 classes of sound events. The dataset consists of 3 subsets: a balanced training set, an unbalanced training set, and an evaluation set. Due to computational limitations, we use the balanced set as the training set and evaluation set for testing. The balanced training set contains 22k samples, and the evaluation set contains 20k samples. The audio recordings were obtained at a sampling rate of 44.1 kHz. For preprocessing, the audio clips are padded or truncated into 10-second recordings. Then, we convert them into log Mel spectrograms, the acoustic features of the audio recordings. Hence, each feature input of $X \in R^{1056\times128}$ corresponds to the log Mel representation of a recording segment with different lengths. 


For training, we designate samples containing specific event classes as positive and those lacking them as negative. For the experiment study, negative audio bags were sampled using different sampling strategies based on a specified Positive-Negative ratio in each training epoch.  We focus on a subset of 40 classes over the entire AudioSet, a subset extracted from the original Audioset. The difference is that AudioSet40 only contains audio recordings with the top 40 classes of events that appear most frequently in real life. Based on this smaller dataset, it would be more efficient for us to implement strategies and compare results. We choose PSLA \cite{gong2021psla}, one of the state-of-the-art methods used in audio classification, as the classification model for the following experiments. Our implementation is based on the PyTorch framework. Hyper-parameters are tuned on our dataset according to PSLA experiments in \cite{gong2021psla}.









\subsubsection{Data Augmentation}
We applied the same data augmentation as \cite{gong2021psla}: MixUp \cite{DBLP:journals/corr/abs-1711-10282} and Time and Frequency masking ~\cite{park19e_interspeech} to improve the performance of the classification on audio. The input spectrogram has the shape of (1056, 128).
Frequency masking is applied to the frequency domain. The mask window is sampled from $f \sim uniform(0,F)$ and the start frequency for masking is drawn from $f_0 \sim uniform(0,128-f)$. Time masking is applied to the time domain so that spectrogram in $[t_0, t_0 + t]$ frames are masked, where $t ~ uniform(0,T)$ and $t_0 \sim uniform(0, 1056-t)$. We use the same parameter $F=48$ and $T=192$ as \cite{gong2021psla}.

Mixup augmentation combines the waveform of two different training audio recordings \(x_i\) and \(x_j\) and their labels \(y_i\) and \(y_j\). The combined waveforms and labels are given by:

\begin{align}
x &= \lambda x_{i} + (1-\lambda) x_{j} \\
y &= \lambda y_{i} + (1-\lambda) y_{j}
\end{align}

Where \(\lambda\) is drawn from a beta distribution:

\begin{equation}
Beta(\alpha, \alpha): \; p(x;\alpha, \alpha) = \frac{x^{\alpha -1}(1-x)^{\alpha -1}}{B(\alpha, \alpha)}
\end{equation}

We set $\alpha=10$ and a mix-up rate 0.5 to achieve better classification performance. We run experiments using Python v3.8.8, Pytorch v1.10.0, and CUDA v11.4 on NVIDIA GeForce RTX 2080Ti. By comparing the results of multiple experiments, we decided to use the following parameters and training strategy: Most of the samples have a length of 10 seconds so we pad or truncate the other audio clips to 10 seconds. We use a batch size of 32. We sample the negative with the sampling strategies and keep all the positive samples in a batch. Weighted batch sampling can improve the performance of classes with a small number of associated samples~\cite{DBLP:journals/corr/abs-2105-05736}. Then, we perform data augmentation on those samples and shuffle them. All the parameters are randomly initialized. Adam optimizer is used with learning rate as $1e^{-3}$ for 30 epochs. Weight decay of $5e^{-7}$ is added to our model to alleviate overfitting. We fix the P-N ratio to 0.5 and run experiments for all Top 40 classes.

\subsubsection{Experiment Result}
Our results are in Figure \ref{fig:audioset_result} as above. Among the sampling strategies experimented with, the gradient embedding method works the best, improving the performance of 28 classes from the random sampling baseline. BADGE takes uncertainty and diversity of the negative bags simultaneously, outperforming random in 26 classes. Gradient-based methods select negative bags that can construct a more efficacious decision boundary. The trade-off between efficiency and efficacy is the main limitation of the gradient-based methods. On average, our code takes about 4 hours to run 30 epochs per class.

\section{Conclusion}
In this paper, we study the importance of sampling negatives in a weak-label learning setting and have experimented with a number of sampling strategies on the CIFAR10 and AudioSet. We find that gradient embedding and BADGE sampling strategies consistently outperform random, margin, and entropy-based sampling strategies. These gradient-based strategies quantify the similarity of the positive bags and the negative ones allowing it to select more informative samples, which leads to improvement in weak label classification. All the proposed sampling strategies have improved in at least 40\% of the classes for AudioSet.


\bibliographystyle{IEEEbib}
\bibliography{strings,refs}

\end{document}